%% file: main.tex
\documentclass[runningheads]{llncs}

\usepackage{eccv}

\usepackage{eccvabbrv}

\usepackage{graphicx}
\usepackage{booktabs}
\usepackage{multirow}
\usepackage{siunitx}
\usepackage{algorithm}
\usepackage{algpseudocode}

\providecommand{\citep}{\cite}
\providecommand{\citet}{\cite}

\usepackage[accsupp]{axessibility}  

\usepackage{hyperref}

\usepackage{orcidlink}

\begin{document}

\title{CSS-BA: Gate-Guided Column Space Search for Bundle Adjustment}

\titlerunning{CSS-BA}

\author{Ayano Kaneda\inst{1} \and
Takafumi Taketomi\inst{2}\orcidlink{0000-0002-5353-0895} \and
Shugo Yamaguchi\inst{1}\orcidlink{0009-0001-6369-3540} \and \\
Shigeo Morishima\inst{1}\orcidlink{0000-0001-8859-6539}}

\authorrunning{A.~Kaneda et al.}

\institute{Waseda University, Tokyo, Japan \and
CyberAgent, Inc., Tokyo, Japan\\
}

\maketitle

\input{sec/0_abstract}
\input{sec/1_intro}
\input{sec/2_relatedwork}
\input{sec/3_method}
\input{sec/4_results}
\input{sec/5_conclusion}

\bibliographystyle{splncs04}
\bibliography{main}

\clearpage
\input{sec/X_suppl}

\end{document}

%% file: sec/0_abstract.tex
\begin{abstract}
Bundle adjustment (BA) remains a critical refinement module for image-based 3D reconstruction and continues to improve geometric accuracy even in learning-based pipelines. However, in low-parallax and near-rotational regimes, classical Schur-based Levenberg--Marquardt (LM) often becomes ill-conditioned and yields unreliable pose and calibration estimates.
We propose Gate-Guided CSS-BA, a solver-side modification of Schur-LM that preserves the classical BA objective and trust-region framework while constraining each update to a geometrically informed low-dimensional subspace. By integrating Column Space Search (CSS) with geometry-aware gating, the method stabilizes the Schur-LM update without altering the estimation problem.
In contrast to keyframe or state-selection approaches, all camera and point parameters remain in the optimization problem; only the update direction is restricted. The method serves as a drop-in replacement for existing BA pipelines.
Experiments on both generic and challenging weak-geometry scenarios show
more stable optimization, improved relative pose accuracy, and competitive calibration behavior while maintaining reprojection quality.
\end{abstract}

%% file: sec/1_intro.tex
\section{Introduction}
\label{sec:intro}
Despite rapid advances in learning-based and foundation-model approaches~\cite{dust3r_cvpr24,MASt3RSfm,wang2025vggt}, bundle adjustment (BA) remains the dominant refinement stage for achieving high geometric accuracy.
In practice, BA often determines the final pose and calibration quality.
However, in low-parallax or near-rotational regimes, the Schur-reduced camera system becomes ill-conditioned.
This is a BA-level issue: image observations constrain bearing directions but weakly separate translation, point depth, and focal calibration when the camera baseline is small.
In a simplified two-view model, the disparity satisfies $d_{\mathrm{disp}}\approx fB/Z$, where $f$ is focal length, $B$ is baseline, and $Z$ is scene depth; since $\partial d_{\mathrm{disp}}/\partial Z\approx -fB/Z^2$, depth sensitivity vanishes as $B\to 0$. This entanglement of translation, depth, and focal length creates weakly constrained directions in the linearized system that, after Schur point elimination, manifest as near-null camera-increment directions, regardless of the specific linear solver used.
Classical Schur-based Levenberg--Marquardt (LM) may converge to solutions with small reprojection error while producing inaccurate relative poses and unstable calibration.
This discrepancy between reprojection accuracy and geometric reliability persists across both classical and learning-based pipelines.

Significant effort has been devoted to improving scalability and numerical efficiency of BA, including square root BA~\cite{Demmel2021RootBA}, matrix-free formulations~\cite{Safari2025SIBA}, and inverse-expansion methods such as Power BA~\cite{Weber2023PoBA} and its variable-projection extensions~\cite{Weber2024PoVar}.
While these approaches enhance conditioning or computational efficiency, they do not directly address the instability of the Schur-LM update under weak geometric conditioning.

We therefore take a complementary solver-side perspective.
Rather than modifying the BA objective or its linearization, we retain the classical Schur-reduced camera system and trust-region framework, and instead restrict each LM step to a geometrically informed low-dimensional search subspace.
In this way, we directly stabilize the update direction while preserving the full objective and variable set, aiming to improve relative pose quality without sacrificing reprojection accuracy.

We refer to the resulting method as \textbf{Gate-Guided CSS-BA}, a selective LM scheme operating entirely within the classical trust-region (TR) framework.
At each iteration, a gated support of informative camera blocks is formed, and Column Space Search (CSS) constructs a compact LM basis from this support.
Camera blocks are ranked using a Schur-consistent predicted reduction score.
A low-dimensional search subspace is then constructed from the camera-block coordinate subspaces spanned by a small high-gain subset of the Schur camera system.

By concentrating updates on well-conditioned directions, the method improves trust-region acceptance and stability in weak geometry while remaining compatible with standard Schur-based BA infrastructures.

Our contributions are summarized as follows:
\begin{itemize}
    \item \textbf{CSS instantiated on Schur-reduced BA.} We adapt Column Space Search to the Schur camera system and construct a compact search basis from gate-guided block statistics without modifying the objective, residuals, or Jacobians.
    \item \textbf{Direction-only restriction with full variables.} Top-$k$ gated blocks are used only to localize the LM step direction; all camera variables remain in the optimization problem.
    \item \textbf{Stability gains in weak geometry.} The method improves relative pose accuracy in low-parallax regimes and can reduce focal-length calibration error under the hardest weak-geometry settings, while preserving reprojection quality and lowering the effective solve dimension.
\end{itemize}

%% file: sec/2_relatedwork.tex
\section{Related Work}
\label{sec:relatedwork}
Bundle adjustment (BA) is the standard nonlinear least-squares refinement in multiview geometry~\citep{Triggs2000BA}. Modern systems rely on Schur-complement reduction and sparse LM solvers, as implemented in widely used libraries such as SBA~\citep{Lourakis2009SBA}, g2o~\citep{Kuemmerle2011g2o}, Ceres~\citep{AgarwalCeres}, and COLMAP~\citep{Schonberger2016COLMAP}.

Prior work on BA has primarily focused on scalability, numerical conditioning, and alternative factorizations. In this section, we review efficiency- and stability-oriented approaches and position our solver-side update strategy within this landscape.

\subsection{Efficiency-oriented methods}

A major line of work improves the scalability of BA by reworking its linear algebra backend. Agarwal et al.~\cite{Agarwal2010BAL} replace sparse Cholesky with preconditioned conjugate gradients on the Schur-reduced system, enabling large-scale Internet reconstructions. Subsequent efforts focus on parallelization and memory efficiency, including multicore acceleration~\cite{Wu2011MCBA}, out-of-core optimization~\cite{Ni2007OutOfCoreBA}, and distributed solvers~\cite{Zhang2017GCC,Ren2022MegBA}.

More recent approaches explore approximate or matrix-free formulations. Zhou et al.~\cite{Zhou2020STBA} reduce computational cost via stochastic approximations, while Safari~\cite{Safari2025SIBA} proposes a matrix-free scheme for shared-intrinsic settings. In parallel, inverse-expansion methods such as Power BA~\cite{Weber2023PoBA} and its variable-projection extension~\cite{Weber2024PoVar} improve efficiency by approximating the Schur complement. Recent GPU-oriented work further accelerates the BA linearization and solve stages~\cite{Han2024EagerBA,InstantSfM2025}.

These works primarily target scalability and conditioning through alternative linear solvers or factorizations.

\subsection{Stability-oriented methods}

A complementary line of work focuses on improving numerical robustness of BA under ill-conditioning and weak initialization.
Square Root BA replaces the standard Schur elimination with a QR-based null-space marginalization, improving conditioning and enabling accurate single-precision solves~\cite{Demmel2021RootBA}.

Variable projection (VarPro) methods analytically eliminate separable parameters to enlarge convergence basins.
Hong et al.~\cite{Hong2016ProjectiveBA,Hong2018pOSE} applied VarPro to BA to obtain initialization-free formulations, and Iglesias et al.~\cite{Iglesias2023expOSE} introduced expOSE, a projective factorization with exponential regularization.
Weber et al.~\cite{Weber2023PoBA,Weber2024PoVar} further extended this direction: PoBA approximates the inverse Schur complement via power-series expansion, while PoVar combines power expansion with VarPro for large-scale initialization-free optimization.

Other recent works explore robustness through probabilistic or calibrated formulations, such as ProBA~\cite{Chui2025ProBA} and calibrated BA without initialization~\cite{Olsson2025InitFreeCalibratedBA}.
As a representative stress case, spherical SfM~\cite{Ventura2025SphericalSfM} highlights the rank-deficiency arising under near-spherical motion, where classical LM can converge to geometrically unreliable solutions.

These approaches improve stability primarily by altering factorizations, reparameterizations, or eliminating subsets of variables.
In contrast, our method retains the classical Schur-reduced system and objective, and instead stabilizes the solver by restricting each LM update to a geometrically informed search subspace.

\subsection{Selective LM under Trust Region}

Our approach is grounded in the trust-region (TR) interpretation of Levenberg--Marquardt (LM).
Under standard assumptions, LM enjoys quadratic local convergence, and with rank deficiency one can still obtain fast local rates under suitable error-bound conditions~\cite{Kanzow2004LMConvexConstraints}.
The trust-region acceptance rule further ensures global descent even when the Jacobian is not locally full rank.

Motivated by this trust-region view, we adopt Column Space Search (CSS)~\cite{Hyde2021SparseSemantic} to construct a selective solver-side update policy within the classical LM framework.
In its original form, CSS scores candidate coordinate directions using residual/Jacobian information, prioritizes data-consistent directions, and solves a reduced subproblem without modifying the objective.

Related lines of work restrict updates by selecting keyframes or states over time (e.g., incremental smoothing, double-window optimization, relative BA, and rolling-shutter BA)~\citep{Kaess2012iSAM2,Strasdat2011DoubleWindow,Sibley2009RBA,Liao2023RSBA}.
Unlike these approaches, which alter the active variable set or measurement support, our method keeps the full BA objective and variable space fixed, and restricts only the solver-side search subspace at each LM iteration.

In our formulation, a geometrically supported block set is identified per iteration, a compact search basis is constructed, and a damped LM subproblem is solved in this subspace before lifting the increment back to the full variable space.
Because acceptance and damping updates follow the same trust-region rule as the baseline LM solver, accepted steps remain governed by the same objective-decrease test while reducing exposure to ill-conditioned directions under weak geometry.

%% file: sec/3_method.tex
\section{Method: Gate-Guided CSS-BA}

\begin{figure*}[t]
\includegraphics[width=1.0\linewidth]{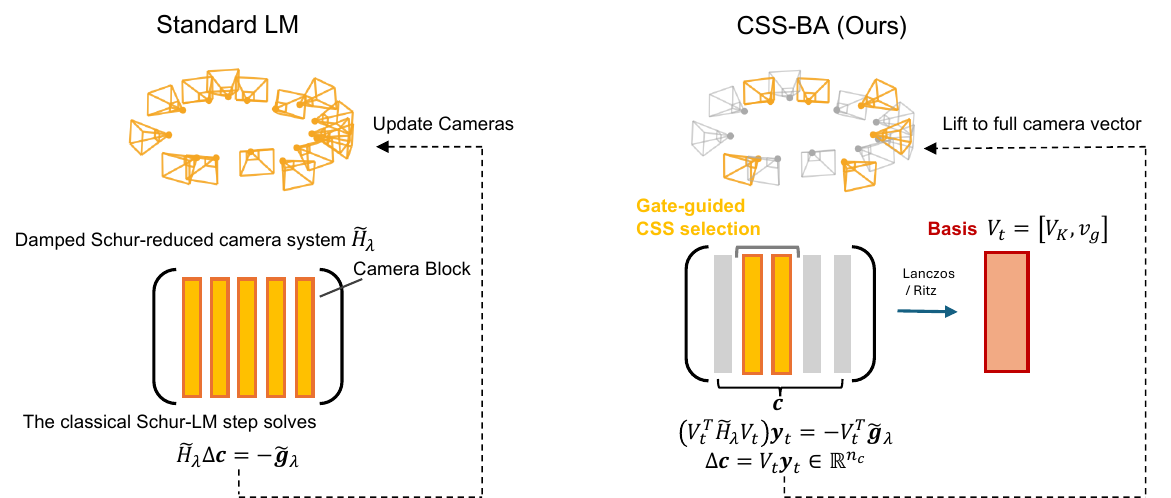}
\caption{Comparison between standard BA and the proposed CSS-BA on the damped Schur camera system.
Standard BA directly solves the full damped Schur system to obtain a full-dimensional camera update.
In contrast, CSS-BA first selects a top-$k$ camera-block support $K_t$, constructs a low-dimensional Lanczos/Ritz basis on the localized Schur operator, and solves the projected LM system in this subspace.
The resulting step is lifted back to the full camera space, so all camera variables remain in the optimization problem, while each iteration restricts the admissible update directions.
The selected blocks determine camera-block \emph{coordinate subspaces} for the basis, not an active-variable mask.}

\label{fig:idea_of_css}
\end{figure*}

\Cref{fig:idea_of_css} provides an overview of the proposed Gate-Guided CSS-BA pipeline.
We design a solver-side Levenberg--Marquardt (LM) scheme tailored to the Schur-reduced bundle adjustment (BA) system.
In classical Schur-LM, full-dimensional camera updates are computed at each iteration.
Under weak geometric conditioning, such as low-parallax or near-rotational motion, the reduced camera system can become ill-conditioned, causing the LM step to move along poorly supported directions despite small reprojection error.

Rather than modifying the BA objective, altering its linearization, or eliminating variables, we retain the standard Schur-reduced formulation and trust-region mechanism.
Instead, we restrict each LM step to a geometrically informed low-dimensional search subspace defined in the Schur-reduced camera increment coordinates.

Unlike keyframe or state-selection methods, the objective and variable set remain unchanged; only the search directions used to compute $\Delta\mathbf c$ are constrained.
The resulting increment is lifted back to the full camera space, so all camera variables remain in the optimization problem, although each iteration restricts the admissible update directions.

\subsection{Preliminaries: Schur-Reduced BA}
We consider the standard bundle adjustment objective
\begin{equation}
E(\mathbf{x}) = \tfrac{1}{2}\sum_{i,j}\|\mathbf{r}_{ij}(\mathbf{c}_i,\mathbf{p}_j)\|_2^2,
\qquad \mathbf{x}=[\mathbf{c};\mathbf{p}],
\end{equation}
where $\mathbf{c}_i$ and $\mathbf{p}_j$ denote camera and point parameters.
Linearization at each LM iteration yields the damped normal equations
\begin{equation}
\bigl(H+\lambda D_{\mathbf x}^{\top}D_{\mathbf x}\bigr)
\Delta\mathbf{x}
=-\mathbf{g},
\qquad
H=J^\top WJ,\quad
\mathbf{g}=J^\top W\mathbf r .
\end{equation}
Here, $\mathbf r$ denotes reprojection errors,
$J$ the Jacobian of the reprojection errors,
$W$ a weighting matrix (e.g., from a robust loss),
$H$ the Gauss--Newton approximation to the Hessian,
$\lambda>0$ the LM damping parameter, and
$D_{\mathbf x}:=\operatorname{blkdiag}(D_c,D_p)$
the diagonal LM scaling matrix for camera and point variables.

Let
\[
H_{cc,\lambda}:=H_{cc}+\lambda D_c^\top D_c,
\qquad
H_{pp,\lambda}:=H_{pp}+\lambda D_p^\top D_p .
\]
Eliminating the point increments gives the damped Schur system
\begin{equation}
\tilde H_\lambda\Delta\mathbf c
=-\tilde{\mathbf g}_\lambda,
\qquad
\tilde H_\lambda
:=H_{cc,\lambda}
-H_{cp}H_{pp,\lambda}^{-1}H_{pc},
\quad
\tilde{\mathbf g}_\lambda
:=\mathbf g_c
-H_{cp}H_{pp,\lambda}^{-1}\mathbf g_p .
\label{eq:css:schur}
\end{equation}

The candidate update $\Delta\mathbf x=[\Delta\mathbf c;\Delta\mathbf p]$ is accepted or rejected using the standard trust-region ratio
\begin{equation}
\rho =
\frac{E(\mathbf{x})-E(\mathbf{x}+\Delta\mathbf{x})}{
-\,\mathbf g^\top\Delta\mathbf x
-\tfrac12\Delta\mathbf x^\top H\Delta\mathbf x}.
\label{eq:rho}
\end{equation}

We retain the baseline LM damping and trust-region update unchanged.
Our modification targets only how $\Delta\mathbf c$ is computed from the damped Schur system in \eqref{eq:css:schur}.

After Schur reduction, the linear solve is expressed over camera increments, and landmark information remains encoded through point elimination.
This reduced system is naturally block-structured by camera blocks.
As discussed in Sec.~\ref{sec:intro}, under low-parallax or near-rotational motion this system becomes ill-conditioned, creating near-null camera-increment directions that cause the LM step to move along poorly supported directions.
Intuitively, small baselines provide weak sensitivity to translation, depth, and focal-length changes; after point elimination, these weak modes appear as near-null directions in the reduced camera system.

\subsection{Column-Space Search as LM Search-Space Design}
\label{sec:subspace_lm}
Column Space Search (CSS)~\cite{Hyde2021SparseSemantic} is a general subspace strategy for nonlinear least squares.
It scores candidate coordinate subspaces and computes an update in the span of a small, data-consistent subset.

In the Schur-reduced BA system, the unknown of the linear solve is the camera increment $\Delta\mathbf c$.
Therefore, we instantiate CSS over camera-block coordinate subspaces of $\Delta\mathbf c$.
At iteration $t$, CSS scores camera blocks and selects a top-$k$ support $K_t$ as described in Sec.~\ref{sec:subspace_build}.
The selected support is then used to construct a basis $V_t$ for a low-dimensional search subspace.

This support should not be interpreted as selecting columns of the Schur matrix or as solving a separate BA problem on selected cameras.
Each selected camera block defines a coordinate subspace of the camera-increment vector.
Equivalently, using the block projector $P_{K_t}$ introduced in Sec.~\ref{sec:subspace_build}, the support localizes the symmetric Schur operator on both sides:
$\tilde H_{\lambda,K_t}=P_{K_t}\tilde H_\lambda P_{K_t}$.

Let $\mathcal U_t\subseteq\mathbb R^{n_c}$ be the search subspace at iteration $t$, where $n_c$ is the total dimension of the camera-parameter vector $\mathbf c$.
Let $V_t\in\mathbb R^{n_c\times r_t}$ be a full column-rank basis of $\mathcal U_t$, constructed as described in Sec.~\ref{sec:subspace_build}.
We restrict only the camera-update direction by parameterizing
$\Delta\mathbf c=V_t\mathbf y_t$.
Substituting this into the damped Schur system in Eq.~\eqref{eq:css:schur} gives the projected LM system
\begin{equation}
\bigl(V_t^\top\tilde H_\lambda V_t\bigr)\mathbf y_t
=
-\,V_t^\top\tilde{\mathbf g}_\lambda,
\qquad
\Delta\mathbf c=V_t\mathbf y_t .
\label{eq:css:projLM}
\end{equation}

After solving for $\Delta\mathbf c$, we recover $\Delta\mathbf p$ by back-substitution using $H_{pp,\lambda}$.
Acceptance and damping updates follow the same trust-region rule in Eq.~\eqref{eq:rho} as the baseline LM solver.

The BA objective, residuals, and variable set remain unchanged.
The increment $\Delta\mathbf c=V_t\mathbf y_t$ is a full-sized camera vector, but its admissible direction is restricted to $\operatorname{range}(V_t)$ at the current iteration.
Thus, $K_t$ is a basis-construction support, not an active-variable mask.

\subsection{CSS Subspace Construction}
\label{sec:subspace_build}

To construct a stable search subspace, we first identify cameras that are geometrically reliable under weak-geometry conditions.
These cameras provide directions that are better conditioned in the Schur-reduced system and are therefore suitable for constructing a stable LM step.

Let $t$ denote the LM iteration index.
We distinguish two camera index sets: $\mathcal S$ is the geometry-aware support, computed once before optimization, and $K_t\subseteq\mathcal S$ is the top-$k$ basis seed selected by CSS scoring at iteration $t$.
Here $k=\min(k_{\max},|\mathcal S|)$, with $k_{\max}=10$ in our experiments.

The set $\mathcal S$ collects cameras that pass geometric reliability tests.

At iteration $t$, CSS scores the camera blocks in $\mathcal S$ and selects $K_t$ to construct the subspace basis; high-gain blocks are those with large local predicted decrease in the current damped Schur model.

\textbf{Gate decides eligibility; CSS decides which eligible blocks are informative.}
The gate forms a reliable support $\mathcal S$ using view connectivity, parallax, and rotational consistency.
CSS then ranks only blocks within $\mathcal S$ by their local Schur-LM predicted decrease and selects the top-$k$ set $K_t$.
Together, gating restricts the candidate pool for basis construction to well-constrained cameras, while CSS selects the most useful eligible blocks for the current LM update.

Importantly, $K_t$ determines basis directions only; it is \emph{not} an active-variable mask on camera parameters.

\subsubsection*{Geometry-aware support $\mathcal S$}

We identify geometrically informative cameras using diagnostics commonly employed in SfM and SLAM keyframe selection, such as parallax magnitude, rotational consistency, and view connectivity.
These quantities detect weakly constrained configurations that often yield ill-conditioned Schur systems.

For camera pose $(R_i,\mathbf t_i)$ with center
\begin{equation}
\mathbf C_i := -R_i^\top \mathbf t_i ,
\end{equation}
we define valid two-view edges
\begin{equation}
(i,j)\in\mathcal E\Longleftrightarrow|\mathcal P_{ij}|\ge\tau_{\mathrm{shared}},
\label{eq:valid_edge}
\end{equation}
where $\mathcal P_{ij}$ denotes the set of shared tracks observed in both cameras $i$ and $j$, with neighborhood $\mathcal N(i)=\{j\mid(i,j)\in\mathcal E\}$.

For each shared track $q\in\mathcal P_{ij}$ with 3D point $\mathbf X_q$, define rays
\begin{equation}
\mathbf u_{iq}=\frac{\mathbf X_q-\mathbf C_i}{\|\mathbf X_q-\mathbf C_i\|},\qquad\mathbf u_{jq}=\frac{\mathbf X_q-\mathbf C_j}{\|\mathbf X_q-\mathbf C_j\|},
\end{equation}
and pairwise median parallax
\begin{equation}
\mathrm{Par}_{ij}=\operatorname{median}_{q\in\mathcal P_{ij}}\arccos\!\big(\operatorname{clamp}(\mathbf u_{iq}^\top\mathbf u_{jq},-1,1)\big).
\end{equation}

We define the parallax-filtered neighborhood
\begin{equation}
\mathcal N_E(i)=\{j\in\mathcal N(i)\mid\mathrm{Par}_{ij}\ge\tau_{\mathrm{par}}^{\mathrm{edge}}\}.
\end{equation}

The camera-level parallax score is the median over parallax-filtered neighboring views,
\begin{equation}
\mathrm{Par}_{i}=\operatorname{median}_{j\in\mathcal N_E(i)}\mathrm{Par}_{ij}.
\end{equation}
If $\mathcal N_E(i)=\emptyset$, we set $\mathrm{Par}_i=0$.
For rotational consistency, let $\bar R_i$ denote the mean rotation of cameras in $\mathcal N(i)$.
We compute the Rotation Agreement (RA) score as the geodesic distance between $R_i$ and $\bar R_i$:

\begin{equation}
\mathrm{RA}_i=\arccos\!\left(\operatorname{clamp}\left(\tfrac{\mathrm{tr}(R_i^\top \bar R_i)-1}{2},-1,1\right)\right).
\label{eq:ra}
\end{equation}

Here $\mathrm{RA}_i=0$ means that $R_i$ is aligned with the average orientation of its neighbors; larger values indicate lower local rotational consistency.
We use RA as a consistency filter, while parallax provides the main baseline-related geometric test.

The geometry-aware support is
\begin{equation}
\mathcal S = \left\{i\ \middle||\mathcal N_E(i)|\ge\tau_{\mathrm{nbr}},\quad\mathrm{RA}_i\le\tau_{\mathrm{ra}},
\quad\mathrm{Par}_i\ge\tau_{\mathrm{par}}^{\mathrm{cam}}\right\}.
\end{equation}

Although similar geometric criteria are often used for keyframe selection in SfM and SLAM, here they serve only to define a stable support for subspace construction.
No variables or measurements are removed from the optimization.

\subsubsection{CSS Camera-Block Scoring for \texorpdfstring{$K_t$}{Kt}}
\label{sec:css_scoring}
At iteration $t$, blocks in $\mathcal S$ are scored on the damped Schur system to select the top-$k$ set $K_t$.
We partition the $n_c$-dimensional camera vector $\mathbf{c} = [\mathbf{c}_1;\ldots;\mathbf{c}_B]$ into $B$ camera blocks, where block $i$ has dimension $d_i$ and $n_c = \sum_{i=1}^B d_i$.
Let $P_i \in \mathbb{R}^{n_c \times d_i}$ embed block $i$ into the full vector.

Let $\tilde{\mathbf g}_{\lambda,i}:=P_i^\top\tilde{\mathbf g}_\lambda$ and $\tilde H_{\lambda,ii}:=P_i^\top\tilde H_\lambda P_i$ denote the corresponding blocks of the damped Schur system.

For each camera $i\in\mathcal S$, CSS computes
\begin{equation}
s_i=\tfrac12\tilde{\mathbf g}_{\lambda,i}^{\top}\tilde H_{\lambda,ii}^{-1}\tilde{\mathbf g}_{\lambda,i},
\label{eq:css:score}
\end{equation}
which is the local predicted decrease under single-block activation for the current damped Schur system.

We choose $K_t\subseteq\mathcal S$ as the top-$k$ indices by descending score.

\subsubsection{Basis Construction from \texorpdfstring{$K_t$}{Kt}}
Given $K_t$, define the block projector
\begin{equation}
P_{K_t} \;=\; \sum_{i\in K_t} P_i P_i^\top,
\label{eq:css:PK}
\end{equation}
and the localized Schur operator

\begin{equation}
\tilde H_{\lambda,K_t}=P_{K_t}\tilde H_\lambda P_{K_t}.
\label{eq:css:HK}
\end{equation}

Thus, top-$k$ selection localizes the operator rather than defining a coordinate-only active-variable set.

\begin{algorithm}[!t]
\caption{Gate-Guided CSS-BA}
\label{alg:csslm}
\begin{algorithmic}[1]
\Require Current state $(\mathbf c,\mathbf p)$; precomputed support $\mathcal S$; options ($k_{\max}$, $m$); LM damping $\lambda$; trust-region parameters

\While{not converged}
  \State Linearize residuals; apply LM damping and Schur $\Rightarrow$ damped reduced $(\tilde H_\lambda,\tilde{\mathbf g}_\lambda)$
  \State $\mathrm{rhs}\leftarrow -\tilde{\mathbf g}_\lambda$

  \Comment{CSS scoring and selection inside the geometry-aware support}
  \ForAll{$i\in\mathcal S$}
    \State $s_i \leftarrow \tfrac12\,\tilde{\mathbf g}_{\lambda,i}^\top\tilde H_{\lambda,ii}^{-1}\tilde{\mathbf g}_{\lambda,i}$
  \EndFor
  \State $k \leftarrow \min(k_{\max}, |\mathcal S|)$ \Comment{fixed top-$k$ budget}
  \State $K_t \leftarrow$ top-$k$ indices in $\mathcal S$ by $s_i$
  \State $P_{K_t} \leftarrow \sum_{i\in K_t} P_i P_i^\top$ \Comment{block projector, $P_i \in \mathbb{R}^{n_c \times d_i}$}
  \State $\tilde H_{\lambda,K_t} \leftarrow P_{K_t}\,\tilde H_\lambda\,P_{K_t}$ \Comment{localized operator}

  \Comment{Low-dimensional basis via Lanczos/Ritz}
  \State Lanczos$(\tilde H_{\lambda,K_t}, m)\Rightarrow (Q_m,T_m)$ \Comment{$Q_m$: Krylov basis; $T_m$: tridiagonal matrix}
  \State $C_k \leftarrow$ top-$k$ eigenvectors of $T_m$
  \State $V_K \leftarrow Q_m C_k$ \Comment{$k$ Ritz vectors}
  \State $z_g \leftarrow (I{-}P_{K_t})(-\tilde{\mathbf g}_\lambda)$
    \If{$\|z_g\|_2>\epsilon_g$}
      \State orthogonalize and normalize $z_g$ against $V_K$ to obtain $v_g$
      \State $V_t \leftarrow [V_K, v_g]$ \Comment{$k+1$ basis vectors}
    \Else
      \State $V_t \leftarrow V_K$ \Comment{$k$ basis vectors}
    \EndIf

  \Comment{Projected LM solve and lifting (full-sized update)}
  \State Solve $\big(V_t^\top\tilde H_\lambda V_t\big)\,y = V_t^\top \mathrm{rhs}$
  \State $\Delta\mathbf c \leftarrow V_t y$ \Comment{full camera increment; not restricted to $K_t$ or $\mathcal S$}
    \State Recover $\Delta\mathbf p$ by back-substitution using $H_{pp,\lambda}$
    \State Evaluate the trial state obtained by applying $(\Delta\mathbf c,\Delta\mathbf p)$ on the manifold
    \State \textit{Additional prediction safeguards and hybrid fallback are described in Sec.~\ref{sec:supp_css_ritz_gate}.}
    \State Compute $\rho$ by Eq.~\eqref{eq:rho}
    \If{$\rho>\eta_{\mathrm{acc}}$}
      \State commit the trial update and decrease damping
    \Else
      \State discard the trial update and increase damping
    \EndIf
\EndWhile
\end{algorithmic}
\end{algorithm}

\textbf{Lanczos/Ritz basis.}
The localized operator $\tilde H_{\lambda,K_t}$ concentrates the Schur curvature on the selected camera blocks.
Lanczos is initialized from the negative reduced gradient restricted to the selected support.

Running $m$-step Lanczos on $\tilde H_{\lambda,K_t}$ generates an orthonormal Krylov basis $Q_m \in \mathbb{R}^{n_c \times m}$ and a tridiagonal matrix $T_m \in \mathbb{R}^{m \times m}$.
We take the top-$k$ eigenvectors $C_k$ of $T_m$ and form
\begin{equation}
V_K = Q_m C_k \in \mathbb{R}^{n_c\times k}.
\label{eq:css:VK}
\end{equation}
These vectors capture the dominant curvature directions within the selected support and provide a compact basis for the LM step before adding the optional complement direction.
This converts the selected camera-block support into smooth joint update directions, rather than optimizing the selected cameras independently.

With the default setting, we also consider one complement direction outside the selected support:
\begin{equation}
z_g=(I-P_{K_t})(-\tilde{\mathbf g}_\lambda).
\label{eq:css:vg}
\end{equation}
If $\|z_g\|_2>\epsilon_g$, we orthogonalize and normalize it against $V_K$ to obtain $v_g$ and set
\begin{equation}
V_t=[V_K,\ v_g],
\label{eq:css:V}
\end{equation}
which has $k+1$ columns. Otherwise, we set $V_t=V_K$, which has $k$ columns.
The complete procedure is summarized in Algorithm~\ref{alg:csslm}.

This basis $V_t$ is plugged into \eqref{eq:css:projLM} to solve the projected LM step and lift a full-sized increment.

\subsubsection{Implementation details.}
\label{sec:implementation}
Our implementation reuses the standard BA components: linearization, LM damping, Schur reduction, and manifold updates.
CSS-BA only changes the camera-step subspace.
In preprocessing, we construct the geometry-aware support set $\mathcal S$ from the diagnostics
$\mathrm{RA}_i$, $|\mathcal N_E(i)|$, and $\mathrm{Par}_i$.

During LM iterations, each step scores blocks in $\mathcal S$ by \eqref{eq:css:score}, selects the top-$k$ set $K_t$, constructs the Ritz basis, solves the projected Schur-LM system, and lifts the full camera increment.

The method uses runtime hyperparameters $(k_{\max},m)$ and gate thresholds
$(\tau_{\mathrm{shared}},\tau_{\mathrm{nbr}},\tau_{\mathrm{ra}},\tau_{\mathrm{par}}^{\mathrm{edge}},\tau_{\mathrm{par}}^{\mathrm{cam}})$.
The concrete values used in our experiments are reported in Sec.~4; implementation-level details of prediction safeguards and hybrid fallback are provided in the supplementary material.
\vspace{-6pt}

%% file: sec/4_results.tex
\section{Experiments}
We evaluate Gate-Guided CSS-BA on two complementary regimes designed to test stability and generality.
First, we consider PhoneSweep~\cite{Ventura2025SphericalSfM}, a low-parallax dataset with near-spherical camera motion, to assess robustness under geometric degeneracy and calibration ambiguity.
Second, we evaluate on Bundle Adjustment in the Large (BAL)~\cite{Agarwal2010BAL}, which contains large-scale Internet-photo reconstructions with more diverse motion, to assess solver behavior beyond near-rotational regimes.

Our implementation is built in C++ on top of the publicly available PoBA codebase~\cite{Weber2023PoBA}, ensuring that Schur reduction, linearization, LM damping, and manifold updates are identical across solvers.
Unless otherwise stated, all methods share the same trust-region strategy and stopping criteria to isolate differences in search-space design.

Experiments were conducted on Ubuntu 22.04.3 LTS (Linux 5.4.0-190-generic) with dual-socket AMD EPYC 7742 CPUs (256 hardware threads), 1TB RAM, and a single NVIDIA A100-SXM4 GPU (40GB, CUDA 12.2).

\subsection{Datasets}
\textbf{PhoneSweep}~\cite{Ventura2025SphericalSfM} contains thirteen real-world sequences with near-spherical camera motion (close to pure rotation, extremely low translational parallax), a regime that induces rank deficiency and calibration ambiguity in BA and provides a challenging stress test for solver stability.
\textbf{BAL}~\cite{Agarwal2010BAL} provides large-scale Internet-photo reconstructions with diverse camera trajectories (ladybug, trafalgar, dubrovnik families), and serves as a complementary generality check under well-conditioned motion.

\subsection{Evaluation Metrics}
For PhoneSweep, we report relative rotation and translation accuracies, RRA@$ \tau$ and RTA@$ \tau$, defined as the percentage of camera pairs whose relative rotation or translation angular error is below the threshold
$\tau\in\{5^\circ,15^\circ,30^\circ\}$; higher is better.
We also report AUC@30 (higher is better) and Absolute Focal Error $\mathrm{AFE}=|\hat f-f_\mathrm{gt}|/f_\mathrm{gt}$ (lower is better).
Formal definitions follow Ventura et al.~\cite{Ventura2025SphericalSfM} and are recalled in the supplementary material.

For BAL, we report per-instance objective-improvement statistics $r_i = f_i^\mathrm{final}/f_i^0$ (smaller is better) and $\Delta_i = \log_{10}(f_i^0/f_i^\mathrm{final})$ (larger is better), aggregated as family-wise means.

\subsection{Experimental Setup}
\noindent
\textbf{Common solver settings.}
We compare three solvers at the bundle adjustment stage: a standard Levenberg--Marquardt baseline (\textbf{Normal-LM}), Power Bundle Adjustment (\textbf{PoBA})~\cite{Weber2023PoBA}, and our \textbf{CSS-BA}.
Normal-LM is the classical Schur-reduced LM solver that solves the full damped system \eqref{eq:css:schur} at each iteration without any subspace restriction.
All solvers operate on the same Schur-reduced camera system.

We use identical trust-region parameters across solvers (initial radius $40.0$, acceptance threshold $\eta=0.1$) and stop when the relative decrease in reprojection cost falls below $10^{-2}$ or after at most $150$ iterations.

Unless otherwise noted, we fix $\tau_{\mathrm{nbr}}=2$ and $\tau_{\mathrm{par}}^{\mathrm{cam}}=2.0$ across datasets and adapt only the rotation gate to the motion regime.

For PoBA, we use $20$ power iterations per linear solve.
For CSS-BA, we use top-$k=10$ with $32$ Lanczos iterations.
At each iteration, support cameras are selected from the geometry-aware support $\mathcal S$ using the CSS block-gain score.

\noindent
\textbf{PhoneSweep protocol.}
We evaluate on the iPhone13Mini and Nexus5X subsets.
Each sequence is initialized using a general SfM pipeline with GLOMAP~\cite{pan2024glomap}.
Due to the near-rotational, low-parallax motion pattern, we use a stricter rotation gate ($\tau_{\mathrm{ra}}=8.0^\circ$).

The original PhoneSweep benchmark assumes shared intrinsics across views.
To isolate solver behavior, our primary comparison among Normal-LM, PoBA, and CSS-BA is conducted without shared intrinsics.
We report SphericalSfM+BA only as a reference baseline, not as a directly matched configuration.

\noindent
\textbf{BAL protocol.}
We use the official BAL problem files and preserve the original camera parameterization, 3D points, and observations.
All solvers are initialized from the same provided state and optimized with identical trust-region parameters and stopping criteria.
Since BAL contains more diverse camera motion than PhoneSweep, we adopt a looser rotation gate ($\tau_{\mathrm{ra}}=20^\circ$) to avoid over-pruning informative cameras.

\sisetup{table-number-alignment=right, table-text-alignment=right}
\begin{table*}[t]
  \centering
  \scriptsize
  \setlength{\tabcolsep}{2pt}
  \caption{Results on the PhoneSweep dataset. SphericalSfM with the shared-intrinsic assumption is used as a reference baseline. $\uparrow$ indicates higher is better and $\downarrow$ lower is better. Compared with Normal-LM and PoBA, CSS-BA substantially improves relative-pose accuracy on both devices; focal calibration is competitive but dataset-dependent.}
  \label{tab:phonesweep-like}
  \begin{tabular}{
    @{} l l
    S[table-format=3.2] S[table-format=3.2] S[table-format=3.2]
    S[table-format=3.2] S[table-format=3.2] S[table-format=3.2]
    S[table-format=3.2]
    S[table-format=3.2]
    @{}
  }
    \toprule
    \multicolumn{2}{c}{} &
    \multicolumn{3}{c}{RRA$\uparrow$} &
    \multicolumn{3}{c}{RTA$\uparrow$} &
    \multicolumn{1}{c}{AUC@30$\uparrow$} &
    \multicolumn{1}{c}{AFE$\downarrow$} \\
    \cmidrule(lr){3-5} \cmidrule(lr){6-8}
    Dataset & Method &
      \multicolumn{1}{c}{@5} & \multicolumn{1}{c}{@15} & \multicolumn{1}{c}{@30} &
      \multicolumn{1}{c}{@5} & \multicolumn{1}{c}{@15} & \multicolumn{1}{c}{@30} &
      \multicolumn{1}{c}{} & \multicolumn{1}{c}{} \\
    \midrule

    \multirow{4}{*}{iPhone13Mini}
      & Normal-LM  & 17.40 & 53.49 & 81.38 & 6.24 & 28.73 & 51.26 & 13.34 &  \textbf{115.56} \\
      & PoBA\cite{Weber2023PoBA} &41.58 & 73.98 & 87.62 & 17.22 & 52.99 & 62.13& 42.62 & 169.43 \\
      & CSS-BA & \textbf{85.79} &  \textbf{86.55} &  \textbf{87.82} &  \textbf{77.24}&  \textbf{86.32} &  \textbf{89.68} &  \textbf{79.97}& 158.01 \\\cmidrule(lr){2-10}
      & Baseline & 100.00 & 100.00 & 100.00 & 86.65 & 98.77 & 99.77& 91.45 & 0.25\\
   \midrule

    \multirow{4}{*}{Nexus5X}
      & Normal-LM  & 8.06& 44.32&76.29& 5.10 & 21.38& 43.32& 12.06 & 265.35\\
      & PoBA\cite{Weber2023PoBA} &14.15 & 66.90 & 88.10 &  16.14 & 52.90 & 75.04 & 33.01 & 121.18 \\
      & CSS-BA  &   \textbf{86.32} &  \textbf{91.09} &  \textbf{95.98} &  \textbf{81.12} &  \textbf{94.79} &  \textbf{98.41} &  \textbf{83.78} &  \textbf{0.89}\\\cmidrule(lr){2-10}
      & Baseline & 100.00 & 100.00 & 100.00 & 83.48 & 96.56 & 98.83 & 90.43&  0.97\\
    \bottomrule
  \end{tabular}
\end{table*}
\sisetup{table-number-alignment=center, table-text-alignment=center}

\begin{table*}[t]
  \centering
  \scriptsize
  \setlength{\tabcolsep}{4pt}
  \caption{BAL family-wise improvement summary. For each problem $i$, we compute
  $r_i=f_i^{\mathrm{final}}/f_i^{0}$ (smaller is better) and
  $\Delta_i=\log_{10}(f_i^{0}/f_i^{\mathrm{final}})$ (larger is better),
  then aggregate within each family (mean over problems).}
  \label{tab:bal-family}
  \begin{tabular}{@{} l c cc cc cc @{}}
    \toprule
    & & \multicolumn{2}{c}{Normal-LM} & \multicolumn{2}{c}{PoBA} & \multicolumn{2}{c}{CSS-BA} \\
    \cmidrule(lr){3-4} \cmidrule(lr){5-6} \cmidrule(lr){7-8}
    Family & \#Probs. & mean $r \downarrow$ & mean $\Delta \uparrow$
           & mean $r \downarrow$ & mean $\Delta \uparrow$
           & mean $r \downarrow$ & mean $\Delta \uparrow$ \\
    \midrule
    ladybug    & 20 & 0.01556 & 1.8713 & 0.016339 & 1.860373 & 0.0172 & 1.8089 \\
    trafalgar  & 14 & 0.00484 & 2.3542 & 0.004707 & 2.338294 & 0.0046 & 2.3400 \\
    dubrovnik  & 16 & 0.00484 & 2.3254 & 0.004907 & 2.319353 & 0.0053 & 2.2919 \\
    \bottomrule
  \end{tabular}
  \vspace{-5.0mm}
\end{table*}

\begin{figure*}[t]
  \centering
  \includegraphics[width=1.0\linewidth]{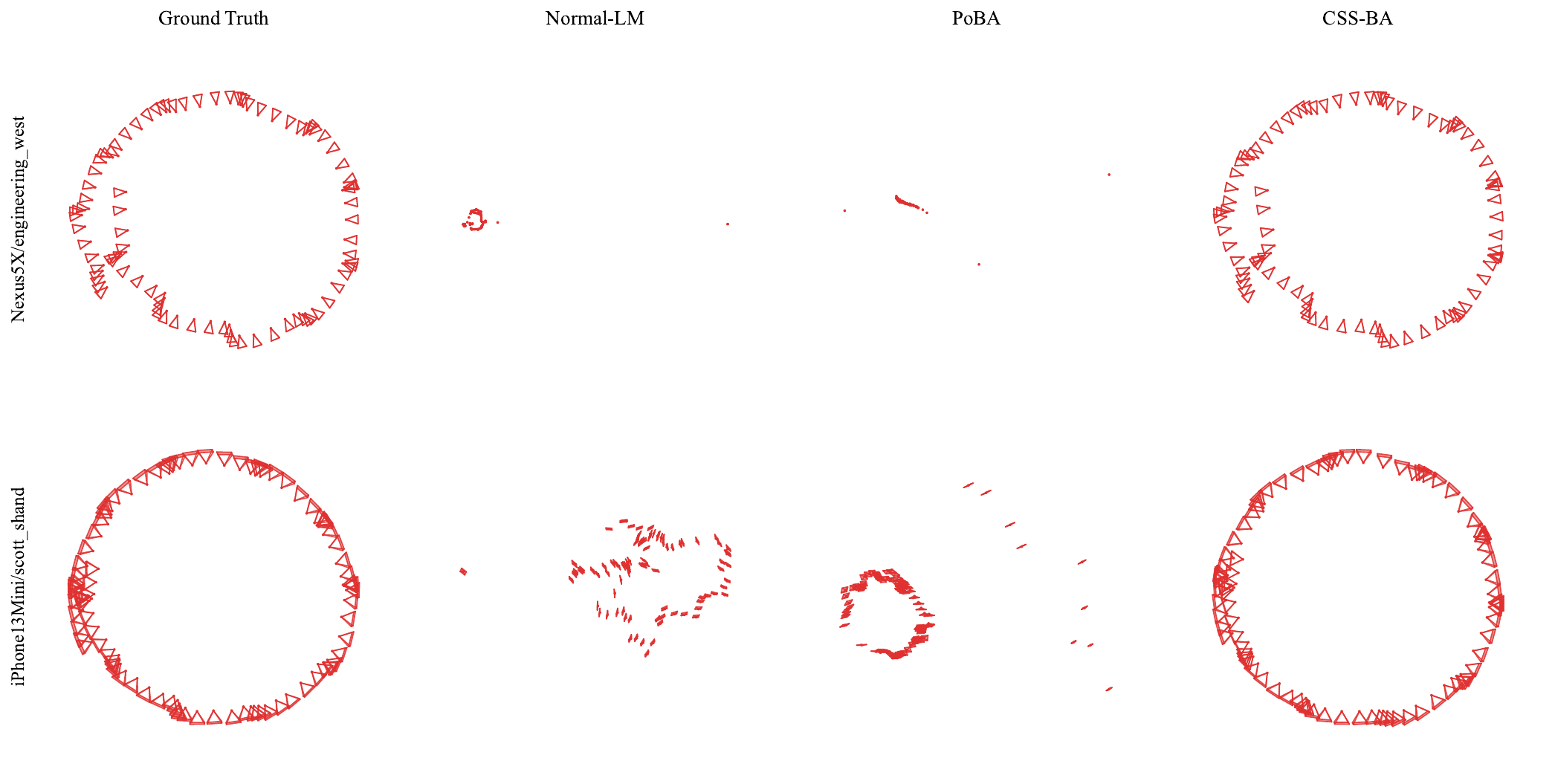}
  \caption{Qualitative comparison of reconstructed camera trajectories on two PhoneSweep sequences: \texttt{Nexus5X/engineering\_west} (top) and \texttt{iPhone13Mini/scott\_shand} (bottom).
Ground-truth camera layouts exhibit the characteristic near-spherical circular trajectory of PhoneSweep.
Normal-LM and PoBA often fail to preserve this geometry, producing collapsed, fragmented, or strongly distorted camera configurations.
In contrast, CSS-BA recovers camera trajectories that remain close to the ground-truth circular layout, indicating substantially improved robustness under low-parallax, near-rotational motion.}
  \label{fig:phonesweep-qualitative}
\end{figure*}

\section{Results}
\subsection{PhoneSweep Results}
Table~\ref{tab:phonesweep-like} compares Normal-LM, PoBA, and CSS-BA on PhoneSweep.
SphericalSfM+BA with shared intrinsics is shown as a reference.

Across both devices, CSS-BA substantially improves relative-pose accuracy over Normal-LM and PoBA.
On iPhone13Mini, AUC@30 rises to $79.97$ from $13.34$ (Normal-LM) and $42.62$ (PoBA).
On Nexus5X, the gain is larger: $83.78$ vs.\ $12.06$ and $33.01$.
On Nexus5X, CSS-BA also reduces AFE from $265.35$/$121.18$ to $0.89$, indicating that the baselines largely fail to recover focal calibration under this extreme degeneracy.

Figure~\ref{fig:phonesweep-qualitative} shows that Normal-LM and PoBA often produce collapsed or fragmented trajectories, while CSS-BA recovers the characteristic near-spherical circular layout.
CSS-BA operates without shared intrinsics or explicit motion constraints, yet approaches the strongly regularized SphericalSfM+BA baseline.

\subsection{BAL Results}
Table~\ref{tab:bal-family} shows that CSS-BA obtains objective reductions comparable to Normal-LM and PoBA on BAL across all three families.
CSS-BA is not consistently the best solver on these less-degenerate problems, but the results suggest that the subspace restriction does not introduce severe degradation in standard large-scale BA settings.
The Venice BAL family is reported in the supplementary material.

\subsection{Ablation}

\noindent\textbf{Gate ablation.}
Table~\ref{tab:gate_ablation} ablates the geometric gate on PhoneSweep (GLOMAP init.).
The gate is a stability filter: it helps most on the harder Nexus5X setting, while iPhone13Mini variants are close due to near-saturation.
The AUC/AFE ordering is not strictly monotone with gate complexity because the update direction is determined by CSS ranking and the Ritz basis, not by the gate alone.

\begin{table}[t]
\centering
\small
\setlength{\tabcolsep}{3.5pt}
\renewcommand{\arraystretch}{0.95}
\caption{Gate ablation on PhoneSweep (GLOMAP initialization). The gate acts as a stability filter; it helps most on the harder Nexus5X setting.}
\label{tab:gate_ablation}
\begin{tabular}{@{}lrrrr@{}}
\toprule
Gate & \multicolumn{2}{c}{iPhone13Mini} & \multicolumn{2}{c}{Nexus5X} \\
\cmidrule(lr){2-3}\cmidrule(l){4-5}
 & AUC@30\,$\uparrow$ & AFE\,$\downarrow$ & AUC@30\,$\uparrow$ & AFE\,$\downarrow$ \\
\midrule
Full gate & 79.97 & 158.01 & 83.78 & 0.89 \\
No gate   & 79.98 & 157.82 & 79.56 & 1.06 \\
RA only   & 79.91 & 157.30 & 79.65 & 0.88 \\
Par only  & 79.94 & 158.25 & 79.61 & 0.94 \\
\bottomrule
\end{tabular}
\end{table}

\noindent\textbf{Top-$k$ sensitivity.}
Table~\ref{tab:topk_sensitivity} reports AUC@30 and AFE for $k \in \{4, 6, 8, 10, 12, 16\}$ on three PhoneSweep sequences under GLOMAP initialization.
Metrics vary little across the range, supporting a fixed default $k=10$ without per-scene tuning.

\begin{table}[t]
\centering
\small
\setlength{\tabcolsep}{2.5pt}
\renewcommand{\arraystretch}{0.95}
\caption{Top-$k$ sensitivity on three PhoneSweep sequences.}
\label{tab:topk_sensitivity}
\begin{tabular}{@{}rcccccc@{}}
\toprule
 & \multicolumn{2}{c}{N5X/eng.}
 & \multicolumn{2}{c}{i13/scott}
 & \multicolumn{2}{c}{i13/plaza} \\
\cmidrule(lr){2-3}\cmidrule(lr){4-5}\cmidrule(l){6-7}
$k$ & AUC@30\,$\uparrow$ & AFE\,$\downarrow$
    & AUC@30\,$\uparrow$ & AFE\,$\downarrow$
    & AUC@30\,$\uparrow$ & AFE\,$\downarrow$ \\
\midrule
4  & 97.747 & 0.503 & 95.365 & 0.601 & 94.327 & 0.692 \\
6  & 97.746 & 0.128 & 95.394 & 0.299 & 93.872 & 0.586 \\
8  & 97.762 & 0.430 & 95.365 & 0.601 & 93.426 & 0.576 \\
10 & 97.753 & 0.447 & 95.363 & 0.601 & 94.249 & 0.654 \\
12 & 97.754 & 0.447 & 95.365 & 0.601 & 93.294 & 0.580 \\
16 & 97.757 & 0.467 & 95.391 & 0.379 & 93.367 & 0.578 \\
\bottomrule
\end{tabular}
\end{table}

\noindent\textbf{Runtime analysis.}
CSS-BA introduces per-iteration overhead from block scoring, Lanczos basis construction, and the projected solve.
Table~\ref{tab:runtime} reports representative wall-clock timings on two PhoneSweep sequences under the same 150-iteration cap.
Aggregating these timed cases gives Normal-LM / PoBA / CSS-BA $\approx$ 6495 / 1930 / 13142\,s, so CSS-BA costs about $2.0\times$ Normal-LM and $6.8\times$ PoBA on this weak-geometry timing subset.
Together with the accuracy gains reported above, this shows that CSS-BA trades throughput for improved pose stability in the targeted ill-conditioned regime.
Calibration gains are strongest on Nexus5X and remain dataset-dependent.

On BAL (Table~\ref{tab:bal-family}), the results show comparable objective improvement; the added subspace construction overhead is therefore mainly justified when weak geometry makes accuracy the primary concern.

\begin{table}[t]
\centering
\small
\setlength{\tabcolsep}{3.5pt}
\caption{Representative wall-clock runtime on two PhoneSweep sequences under the 150-iteration cap.}
\label{tab:runtime}
\begin{tabular}{@{}lrrrrr@{}}
\toprule
Case & Normal-LM & PoBA & CSS-BA & vs. Normal & vs. PoBA \\
\midrule
Seq. A & 2306\,s & 683\,s  & 4478\,s & $1.94\times$ & $6.56\times$ \\
Seq. B & 4189\,s & 1247\,s & 8664\,s & $2.07\times$ & $6.95\times$ \\
Total  & 6495\,s & 1930\,s & 13142\,s & $2.02\times$ & $6.81\times$ \\
\bottomrule
\end{tabular}
\end{table}

%% file: sec/5_conclusion.tex
\section{Conclusion and Limitations}

We introduced Gate-Guided CSS-BA, a solver-side Column Space Search variant of Schur-LM tailored to bundle adjustment under weak geometric conditioning.
Operating on the Schur-reduced camera system, the method combines geometry-aware support selection with a subspace-restricted update while preserving the original objective, variable set, and trust-region mechanism.

On near-spherical, low-parallax PhoneSweep sequences, CSS-BA improves
relative pose accuracy over standard LM and PoBA and, in the Nexus5X case,
substantially reduces focal-length error.
It approaches the strongly regularized SphericalSfM+BA reference in relative-pose
and trajectory-layout metrics without shared intrinsics or explicit motion priors.

Because CSS-BA only changes the LM update subspace, it is compatible with standard robust estimators and outlier handling.
Its main benefit is in ill-conditioned regimes where classical Schur-LM is prone to poorly supported updates; in well-conditioned cases, gains are more modest and subspace construction adds overhead.
Adaptive gating and subspace sizing remain future work.

%% file: sec/X_suppl.tex
\appendix
\renewcommand{\theHsection}{supp.\thesection}
\renewcommand{\theHequation}{supp.\thesection.\arabic{equation}}
\setcounter{figure}{0}
\setcounter{table}{0}
\renewcommand{\thefigure}{A.\arabic{figure}}
\renewcommand{\thetable}{A.\arabic{table}}
\renewcommand{\theHfigure}{supp.fig.\arabic{figure}}
\renewcommand{\theHtable}{supp.tab.\arabic{table}}

\title{Supplementary Material for ``CSS-BA: Gate-Guided Column Space Search for Bundle Adjustment''}
\author{Ayano Kaneda\inst{1} \and
Takafumi Taketomi\inst{2} \and
Shugo Yamaguchi\inst{1} \and \\
Shigeo Morishima\inst{1}}

\institute{Waseda University, Tokyo, Japan \and
CyberAgent, Inc., Tokyo, Japan\\
}

\titlerunning{CSS-BA}
\authorrunning{A.~Kaneda et al.}

\maketitle

\noindent
This supplementary material provides evaluation metric definitions, additional implementation details, robustness checks, and failure-case analyses.

\section{Evaluation Metric Definitions}

\noindent\textbf{Relative pose errors.}
For a camera pair $(i,k)$, let $R_{ik}=R_kR_i^\top$ be the ground-truth relative rotation and $\mathbf t_{ik}=R_i^\top(\mathbf t_k-\mathbf t_i)/\|R_i^\top(\mathbf t_k-\mathbf t_i)\|$ be the ground-truth relative translation direction.
We first define the angular errors
\begin{equation}
\begin{aligned}
e_R(i,k)&=\arccos\!\left(\operatorname{clamp}\left(\frac{\mathrm{tr}(R_{ik}^\top\hat R_{ik})-1}{2},-1,1\right)\right),\\
e_T(i,k)&=\arccos\!\left(\operatorname{clamp}\left(\mathbf t_{ik}^\top\hat{\mathbf t}_{ik},-1,1\right)\right).
\end{aligned}
\end{equation}

\noindent\textbf{RRA / RTA.}
For a threshold $\tau$, the reported relative rotation and translation accuracies are
\begin{equation}
\mathrm{RRA}@\tau=\frac{100}{|\mathcal P|}\sum_{(i,k)\in\mathcal P}\mathbf 1[e_R(i,k)\le\tau],\quad
\mathrm{RTA}@\tau=\frac{100}{|\mathcal P|}\sum_{(i,k)\in\mathcal P}\mathbf 1[e_T(i,k)\le\tau].
\end{equation}

\noindent\textbf{AUC@30.}
For each threshold $\tau$, define the joint pose accuracy
\begin{equation}
\mathrm{Acc}(\tau)=\frac{100}{|\mathcal P|}\sum_{(i,k)\in\mathcal P}\mathbf 1[e_R(i,k)\le\tau\wedge e_T(i,k)\le\tau].
\end{equation}
AUC@30 is the normalized area under $\mathrm{Acc}(\tau)$ over $\tau\in[0,30^\circ]$.

\noindent\textbf{AFE.}
We define the absolute focal error as $\mathrm{AFE}=|\hat f-f_{\mathrm{gt}}|/f_{\mathrm{gt}}$, following Ventura et al.~\cite{Ventura2025SphericalSfM}.

\section{PhoneSweep Results with SphericalSfM Initialization}
In the main paper, we evaluate CSS-BA on PhoneSweep using a general SfM initialization obtained with GLOMAP.
As a complementary experiment, we additionally evaluate a spherical-motion-specific initialization produced by SphericalSfM~\cite{Ventura2025SphericalSfM} before bundle adjustment.
This setting is better aligned with the near-spherical motion of PhoneSweep and allows us to test whether the proposed solver remains effective under a stronger motion-consistent initialization.

For this complementary SphericalSfM-initialized experiment, we use the following device-specific gate thresholds to account for the different PhoneSweep camera characteristics.

For Nexus5X, we use $\tau_{\mathrm{nbr}}=2$, $\tau_{\mathrm{par}}^{\mathrm{cam}}=1.8$, and $\tau_{\mathrm{ra}}=7.1^\circ$.
For iPhone13Mini, we use $\tau_{\mathrm{nbr}}=2$, $\tau_{\mathrm{par}}^{\mathrm{cam}}=0.8$, and $\tau_{\mathrm{ra}}=7.1^\circ$.
All other settings, including the top-$k$ budget $k_{\max}=10$ and the Lanczos iteration count $m=32$, remain unchanged.

\textbf{Quantitative results.}
\Cref{tab:sphericalSfM-init} summarizes the quantitative results.

CSS-BA again improves relative-pose accuracy over Normal-LM and PoBA on both devices and reduces AFE among the three BA solvers.
The SphericalSfM+BA reference remains stronger in focal calibration because it uses the shared-intrinsic assumption.
These gains are obtained without the shared-intrinsic assumption used by the SphericalSfM+BA reference baseline, so the comparison still reflects a more general bundle-adjustment setting.

At the same time, the translation results suggest a more nuanced picture than rotation alone.
For both devices, CSS-BA approaches the reference baseline much more closely at coarse angular tolerances (e.g., RTA@30) than at stricter ones (e.g., RTA@5), indicating that the translational geometry is largely recovered at a coarse level while fine-grained translation-direction accuracy remains harder to obtain.
This behavior is consistent with the extremely weak-parallax nature of PhoneSweep.

\begin{figure*}[t]
  \centering
  \includegraphics[width=1.0\linewidth]{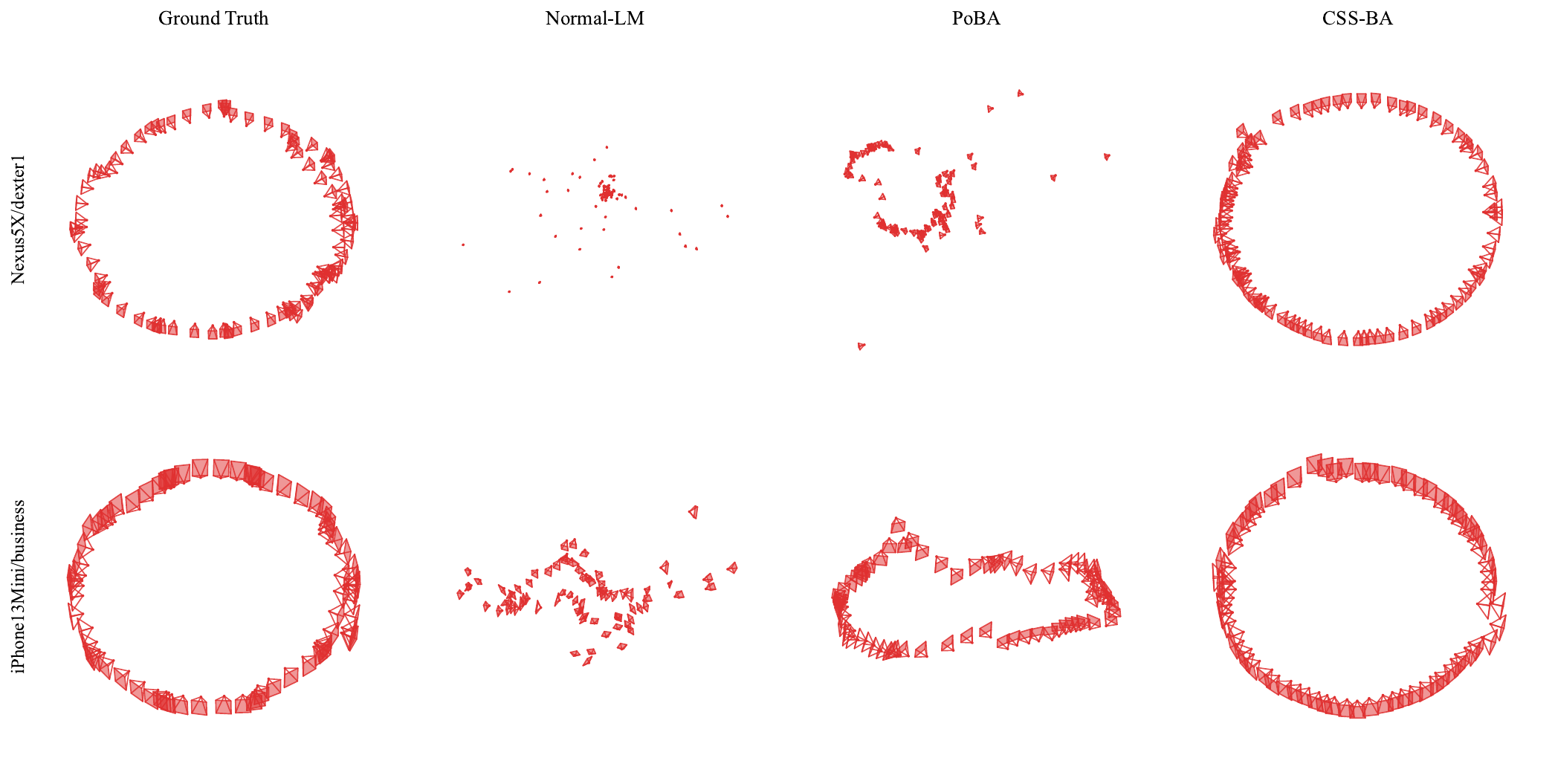}
  \caption{Qualitative comparison of reconstructed camera trajectories on two PhoneSweep sequences under SphericalSfM initialization.
Ground-truth camera layouts exhibit the characteristic near-spherical circular trajectory of PhoneSweep.
Normal-LM and PoBA often fail to preserve this geometry.
In contrast, CSS-BA recovers outward-facing trajectories that remain close to the ground-truth circular layout.}
  \label{fig:visualization_of_cameras}
\end{figure*}

\textbf{Qualitative results.}
\Cref{fig:visualization_of_cameras} provides a qualitative comparison on representative PhoneSweep sequences under SphericalSfM initialization.
CSS-BA recovers outward-facing trajectories that remain close to the ground-truth circular layout.

\sisetup{table-number-alignment=right, table-text-alignment=right}
\begin{table*}[t]
  \centering
  \scriptsize
  \setlength{\tabcolsep}{3pt}
  \caption{Results on PhoneSweep with SphericalSfM initialization. SphericalSfM+BA with the shared-intrinsic assumption is shown only as a reference because it uses a different calibration setting. $\uparrow$ indicates higher is better and $\downarrow$ lower is better. Among the three BA solvers evaluated without shared intrinsics, CSS-BA improves relative-pose accuracy and reduces AFE on both devices.}
  \label{tab:sphericalSfM-init}
  \begin{tabular}{
    @{} l l
    S[table-format=3.2] S[table-format=3.2] S[table-format=3.2]
    S[table-format=3.2] S[table-format=3.2] S[table-format=3.2]
    S[table-format=3.2]
    S[table-format=3.2]
    @{}
  }
    \toprule
    \multicolumn{2}{c}{} &
    \multicolumn{3}{c}{RRA$\uparrow$} &
    \multicolumn{3}{c}{RTA$\uparrow$} &
    \multicolumn{1}{c}{AUC@30$\uparrow$} &
    \multicolumn{1}{c}{AFE$\downarrow$} \\
    \cmidrule(lr){3-5} \cmidrule(lr){6-8}
    Dataset & Method &
      \multicolumn{1}{c}{@5} & \multicolumn{1}{c}{@15} & \multicolumn{1}{c}{@30} &
      \multicolumn{1}{c}{@5} & \multicolumn{1}{c}{@15} & \multicolumn{1}{c}{@30} &
      \multicolumn{1}{c}{} & \multicolumn{1}{c}{} \\
    \midrule

    \multirow{4}{*}{iPhone13Mini}
      & Normal-LM  & 4.36 & 31.03 & 65.41 & 4.80 & 27.52 & 54.68 & 13.04 & 49.14 \\
      & PoBA\cite{Weber2023PoBA} & 42.39& 81.32& 92.97 & 25.61&36.08&76.44& 58.21 &8.01\\
      & CSS-BA & \textbf{93.18} & \textbf{100} & \textbf{100} & \textbf{57.62} & \textbf{89.50} & \textbf{95.72} & \textbf{78.66} & \textbf{1.07} \\\cmidrule(lr){2-10}
      & Reference & 100.00 & 100.00 & 100.00 & 86.65 & 98.77 & 99.77 & 91.45 & 0.25\\
   \midrule

    \multirow{4}{*}{Nexus5X}
      & Normal-LM   &4.10 & 25.53 & 56.27 & 2.98 & 15.16 & 33.40 & 5.12 & 83.92\\
      & PoBA\cite{Weber2023PoBA}  & 55.54& 75.36 & 86.81&10.72 &	23.72 &	52.70 & 33.29 & 18.01\\
      & CSS-BA  & \textbf{87.72} & \textbf{100} & \textbf{100} & \textbf{39.44} & \textbf{73.80} & \textbf{81.24} & \textbf{63.07} & \textbf{5.21}\\\cmidrule(lr){2-10}
      & Reference & 100.00 & 100.00 & 100.00 & 83.48 & 96.56 & 98.83 & 90.43 & 0.97\\
    \bottomrule
  \end{tabular}
\end{table*}

\section{Additional BAL Results on the Venice Family}

To further assess solver behavior beyond the BAL families reported in the main paper, we additionally evaluate on the Venice family, which contains larger problem instances.

\Cref{tab:venice-specific} summarizes the results on a fixed subset of 12 Venice problems.
CSS-BA remains competitive with Normal-LM and PoBA in this larger-scale setting.
Although the gains are smaller than in the strongly degenerate PhoneSweep regime, the results indicate that the proposed solver does not break down on larger BAL instances and remains comparable to conventional solvers under more general geometry.

\begin{table}[t]
\centering
\small
\caption{Results on 12 selected problems from the Venice BAL family.
We report the same objective-improvement statistics as in the main paper: mean $r$ ($\downarrow$) and mean $\Delta$ ($\uparrow$).
CSS-BA remains competitive with Normal-LM and PoBA on these larger-scale BAL instances.}
\label{tab:venice-specific}
\begin{tabular}{@{} l c c @{}}
\toprule
Method & mean $r \downarrow$ & mean $\Delta \uparrow$ \\
\midrule
Normal-LM & 0.00988 & 2.0428 \\
PoBA & 0.00993 & 2.0414 \\
CSS-BA  & 0.00992 & 2.0419 \\
\bottomrule
\end{tabular}
\end{table}

\section{Failure Analysis of GLOMAP Initialization on PhoneSweep}
\label{sec:rationale}

\begin{figure}[t]
  \includegraphics[width=0.9\linewidth]{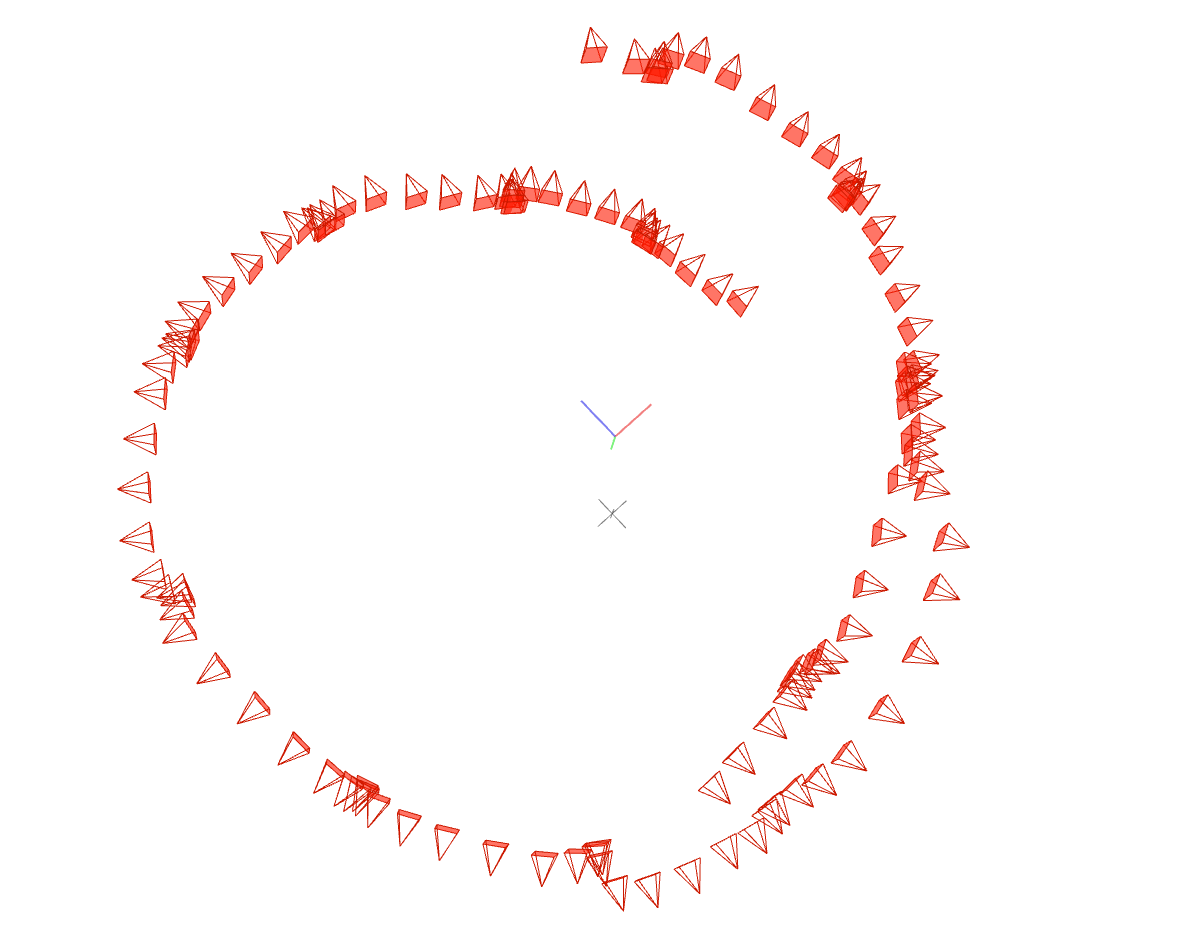}
  \caption{GLOMAP failure on PhoneSweep (Nexus5X baker4): an outward-looking sweep is initialized as inward-facing, producing a globally flipped rig that was not corrected by the tested downstream BA solvers.}
  \label{fig:glomap_failure_example}
\end{figure}

To better understand the PhoneSweep results under GLOMAP initialization reported in the main paper, we inspect representative failure cases of the initialization itself.
As illustrated in \cref{fig:glomap_failure_example}, some outward-looking sequences are incorrectly initialized as inward-facing, indicating that the initializer can produce a globally flipped camera rig before any downstream bundle adjustment is applied.

This observation suggests that part of the degradation in the GLOMAP-initialized setting originates from the initializer rather than from the downstream BA solver.
To isolate this effect, we additionally report a filtered analysis in which sequences with clear outward/inward inversion failures are excluded.
The resulting averages are shown in \cref{tab:phonesweep-except-inversion}.

In this filtered setting, CSS-BA still matches or surpasses Normal-LM and PoBA across the reported pose and calibration metrics.

\begin{table*}[t]
  \centering
  \scriptsize
  \setlength{\tabcolsep}{3pt}
  \caption{PhoneSweep results under GLOMAP initialization after excluding sequences with clear outward/inward inversion failures.
  SphericalSfM with shared intrinsics is shown as a reference baseline.
  $\uparrow$ indicates higher is better and $\downarrow$ lower is better.
  CSS-BA remains at least comparable to Normal-LM and PoBA across the reported metrics in this filtered diagnostic analysis.}
  \label{tab:phonesweep-except-inversion}
  \begin{tabular}{
    @{} l l
    S[table-format=3.2] S[table-format=3.2] S[table-format=3.2]
    S[table-format=3.2] S[table-format=3.2] S[table-format=3.2]
    S[table-format=3.2]
    S[table-format=3.2]
    @{}
  }
    \toprule
    \multicolumn{2}{c}{} &
    \multicolumn{3}{c}{RRA$\uparrow$} &
    \multicolumn{3}{c}{RTA$\uparrow$} &
    \multicolumn{1}{c}{AUC@30$\uparrow$} &
    \multicolumn{1}{c}{AFE$\downarrow$} \\
    \cmidrule(lr){3-5} \cmidrule(lr){6-8}
     Dataset & Method &
      \multicolumn{1}{c}{@5} & \multicolumn{1}{c}{@15} & \multicolumn{1}{c}{@30} &
      \multicolumn{1}{c}{@5} & \multicolumn{1}{c}{@15} & \multicolumn{1}{c}{@30} &
      \multicolumn{1}{c}{} & \multicolumn{1}{c}{} \\
    \midrule

    \multirow{3}{*}{iPhone13Mini}
      & Normal-LM  & 4.536 & 31.81 & 65.56 & 4.820 & 26.66 & 53.51 & 12.85 & 51.51 \\
      & PoBA\cite{Weber2023PoBA} & 45.68 & 78.32 & 89.69 & 20.77 & 58.42 & 68.17 & 47.46 & 174.98 \\
      & CSS-BA & \textbf{100.00} & \textbf{100.00} & \textbf{100.00} & \textbf{89.01} & \textbf{98.71} & \textbf{99.74} & \textbf{93.09} & \textbf{0.65} \\\cmidrule(lr){2-10}
    \multirow{3}{*}{Nexus5X}
      & Normal-LM & 4.626 & 26.00 & 55.76 & 3.501 & 17.58 & 38.05 & 5.770 & 81.49 \\
      & PoBA\cite{Weber2023PoBA} & 16.28 & 76.60 & 95.66 & 17.12 & 54.84 & 76.01 & 37.70 & 132.62 \\
      & CSS-BA & \textbf{100.00} & \textbf{100.00} & \textbf{100.00} & \textbf{91.91} & \textbf{98.39} & \textbf{99.64} & \textbf{94.61} & \textbf{0.40}\\
    \bottomrule
  \end{tabular}
\end{table*}

\section{Additional Implementation Details}
\label{sec:supp_css_ritz_gate}
This section provides implementation details that are omitted from the main paper for brevity.
Unless otherwise noted, the runtime uses a support size of $k=10$ and $32$ Lanczos iterations for the Ritz subspace construction.
At each iteration, the solver scores blocks in the geometry-aware support $\mathcal S$ and selects the top-$k$ support $K_t$ used to build the projected search subspace.

Unless otherwise noted, the runtime includes prediction safeguards and a hybrid fallback.
These mechanisms are triggered only when the projected CSS step fails to provide a sufficiently reliable predicted decrease, and therefore do not affect ordinary successful iterations.
Experiment-specific exceptions are stated in the corresponding sections.

Unless otherwise noted, experiments use the default gate configuration described in the main paper, and any experiment-specific threshold changes are stated explicitly in the corresponding section.

The gate restricts only which camera blocks are eligible for scoring and selection into the CSS subspace; it does not directly constrain the final search direction or determine which variables are allowed to move. After the projected solve, the recovered camera increment remains full-sized, so camera blocks outside the gate or outside the selected support can still receive nonzero updates.
